\documentclass[runningheads]{llncs}

\usepackage[dvipsnames]{xcolor}
\usepackage{multirow} 
\usepackage{amsmath} 
\usepackage{bm} 
\usepackage{bbold}
\usepackage{booktabs}
\usepackage{tikz}
\usetikzlibrary {bending} 
\usetikzlibrary{positioning}
\usetikzlibrary{arrows.meta,decorations, decorations.text,}
\usetikzlibrary{patterns}
\usepackage{hyperref}
\usepackage{relsize}

\usepackage{pgfplots,pgfplotstable}

\usepackage[T1]{fontenc}
\usepackage{graphicx}
\begin{document}
\title{Reevaluation of Inductive Link Prediction}
%
%
\author{Simon Ott\inst{1,2}\orcidID{0000-0003-1926-6899} \and
Christian Meilicke\inst{2}\orcidID{0000-0002-0198-5396} \and
Heiner Stuckenschmidt\inst{2}\orcidID{0000-0002-0209-3859}}
\authorrunning{S. Ott et al.}
%
\institute{AIT Austrian Institute of Technology GmbH, Austria\\
\and
University of Mannheim, Germany}
\maketitle              
\begin{abstract}
Within this paper, we show that the evaluation protocol currently used for inductive link prediction is heavily flawed as it relies on ranking the true entity in a small set of randomly sampled negative entities. Due to the limited size of the set of negatives, a simple rule-based baseline can achieve state-of-the-art results, which simply ranks entities higher based on the validity of their type. As a consequence of these insights, we reevaluate current approaches for inductive link prediction on several benchmarks using the link prediction protocol usually applied to the transductive setting. As some inductive methods suffer from scalability issues when evaluated in this setting, we propose and apply additionally an improved sampling protocol, which does not suffer from the problem mentioned above. The results of our evaluation differ drastically from the results reported in so far.

\end{abstract}
\section{Introduction}
\label{sec:intro}

Knowledge graphs are commonly used to store knowledge in a structured format. However, even well-maintained, large-scale knowledge graphs such as Freebase~\cite{bollacker2008freebase}, DBPedia~\cite{auer2007dbpedia} or the Google Knowledge Graph~\cite{noy2019industry} are notoriously incomplete~\cite{dong2014knowledge}, which limits their usefulness. Given the knowledge already encoded in the graph, some of the missing facts can be entailed or are rather likely due to the probabilistic regularities inherent in the graph. The automated task of predicting missing facts in a knowledge graph without using external knowledge is known as link prediction or knowledge graph completion.

Over time many different approaches were proposed to address this task. Some of the most prominent ones, such as TransE~\cite{bordes2013translating}, Complex~\cite{trouillon2016complex}, ConvE~\cite{dettmers2018convolutional}, and RotateE~\cite{sun2019rotate} are based on embedding entities (constants) and relations (predicates) in a vector space. These embeddings are learned as a solution to an optimization problem which is defined by the triples in the graph and a specific scoring function which determines the likelihood of a triple being correct. Learnt vectors are then applied to the scoring function to entail missing triples.
While these models have proven to perform well in some established evaluation datasets~\cite{rossi2021knowledge}, they can only make predictions about entities that were already seen during training. If no embedding for an entity has been learned, it is impossible to compute a score for this entity. Nevertheless, these models have been dominating knowledge base completion for a long time. This is related to the fact that the entities of the evaluation test sets of the most prominent benchmark datasets, such as FB15k-237~\cite{toutanova2015observed} and WN18RR~\cite{dettmers2018convolutional}, are subsets of the entities appearing in the training set. Such a scenario is referred to as \textbf{transductive} link prediction. However, there might be link prediction applications where it is not realistic to assume that a transductive setting is given.

A scenario, where the training and test set does not overlap is referred to as \textbf{inductive} link prediction~\cite{yang,grail}. The inductive setting has received lots of attention and many different approaches have been proposed~\cite{nodepiece,conglr,indigo,compile,grail,snri,redgnn,astarnet,nbfnet}. Within this work we are concerned with a flaw of the data sets and evaluation principles that are usually applied in the context of inductive link prediction. Both transductive and inductive link prediction evaluation are based on computing rankings of candidates for a given completion query, such as \textit{married(john, ?)}. The positions of the correct candidates in the rankings determine the evaluation result. Surprisingly, there is a major difference in the evaluation protocol not related to the difference between the inductive and the transductive setting: In the transductive setting usually the correct entity is ranked within all entities of the dataset. Contrary to this, in the inductive setting the correct entity is ranked only within a relatively small randomly drawn subset of all candidates. 


We argue that the results obtained by the random sampling evaluation do not measure predictive accuracy but the capability to distinguish between a potentially meaningful candidate and a nonsensical. Suppose we create a random sample of 50 entities from a  knowledge graph such as DBpedia or Freebase. This sample will contain entities of widely varying types, such as organisations, locations, languages, persons, currencies or other. How many entities are persons that might be married to someone? The answer is probably a relatively small number. Thus, it is obviously easy to rank the correct candidate at a top position for our example query given a set of 50 randomly chosen candidates.

Within this paper we analyse the problems of the random sampling evaluation protocol, which has become the de-facto standard for the inductive setting. We propose a simple rule-based baseline which is designed to solve the task to distinguish between a potentially meaningful candidate, that has an appropriate type, and nonsensical candidates. We show in a rich set of experiments that this baseline outperforms many other approaches when applying the random sampling evaluation protocol. This shows that this evaluation protocol is not really measuring predictive accuracy. 

We propose, instead, to rank the correct entity within all possible candidates. If this is not possible due to scalability issues, we release a collection of type-matched negatives (TMN) for each benchmark dataset to rank the candidate within a set of entities that have at least an appropriate type. We perform experiments using both approaches with 13 current state-of-the-art methods. Our experiments prove that the random sampling evaluation yields misleading results and that results obtained by a reasonable evaluation protocol differ significantly. We release code, evaluation datasets, and additional results at \href{https://github.com/nomisto/inductiveeval}{https://github.com/nomisto/inductiveeval}.

\section{Preliminaries}


\subsection{Link Prediction}
\label{sub:kg}

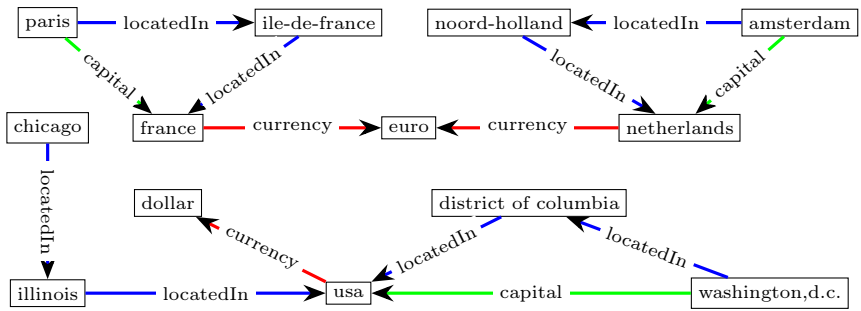
\begin{figure}[b!]
\caption{Example KG of cities, counties, countries and currencies. Different colors represent different relations.}
\label{KG}
\medskip
\centering
\scalebox{1.0}{
    \begin{tikzpicture}[scale=4.0]
    \scriptsize
    \begin{scope}[every node/.style={rectangle,draw}]
        \node (paris) at (0,0) {paris};
        \node (amsterdam) at (2.50, 0) {amsterdam};
        \node (chicago) at (0.0, -0.35) {chicago};
        \node (washington-d-c) at (2.4, -0.9) {washington,d.c.};
    \end{scope}
    \begin{scope}[every node/.style={rectangle,draw}]
        \node (ile-de-france) at (0.9, 0.0) {ile-de-france};
        \node (noord-holland) at (1.5, 0.0) {noord-holland};
        \node (illinois) at (0.0, -0.9) {illinois};
        \node (district of columbia) at (1.6, -0.6) {district of columbia};
    \end{scope}
    \begin{scope}[every node/.style={rectangle,draw}]
        \node (france) at (0.4, -0.35) {france};
        \node (euro) at (1.2, -0.35) {euro};
        \node (usa) at (1.0, -0.9) {usa};
    \end{scope}
    \begin{scope}[every node/.style={rectangle,draw}]
        \node (netherlands) at (2.1, -0.35) {netherlands};
        \node (dollar) at (0.4, -0.6) {dollar};
    \end{scope}
    
    \begin{scope}[>={Stealth[black]},every node/.style={fill=white},every edge/.style={draw=green,very thick}]
        \path [->] (paris) edge[bend right=0] node[sloped] {capital} (france);
        \path [->] (washington-d-c) edge[bend left=0] node[sloped] {capital} (usa);
        \path [->] (amsterdam) edge[bend left=0] node[sloped] {capital} (netherlands);
    \end{scope}

    \begin{scope}[>={Stealth[black]},every node/.style={fill=white},every edge/.style={draw=blue,very thick}]
        \path [->] (paris) edge[bend left=0] node[sloped] {locatedIn} (ile-de-france);
        \path [->] (ile-de-france) edge[bend left=0] node[sloped] {locatedIn} (france);
        \path [->] (amsterdam) edge[bend right=0] node[sloped] {locatedIn} (noord-holland);
        \path [->] (noord-holland) edge[bend right=0] node[sloped] {locatedIn} (netherlands);
        \path [->] (chicago) edge[bend left=0] node[sloped] {locatedIn} (illinois);
        \path [->] (illinois) edge[bend right=0] node[sloped] {locatedIn} (usa);
        \path [->] (washington-d-c) edge[bend right=0] node[sloped] {locatedIn} (district of columbia);
        \path [->] (district of columbia) edge[bend left=0] node[sloped] {locatedIn} (usa);
    \end{scope}

    \begin{scope}[>={Stealth[black]},every node/.style={fill=white},every edge/.style={draw=red,very thick}]
        \path [->] (netherlands) edge[bend left=0] node[sloped] {currency} (euro);
        \path [->] (france) edge[bend right=0] node[sloped] {currency} (euro);
        \path [->] (usa) edge[bend right=0] node[sloped] {currency} (dollar);
    \end{scope}
    
\end{tikzpicture}
}
\end{figure}

A knowledge graph (KG) $\mathcal{G} = (\mathcal{E}, \mathcal{R}, \mathcal{T})$ is a heterogeneous directed multigraph consisting of triples $\mathcal{T}$, a set of entities $\mathcal{E}$ and a set of relations $\mathcal{R}$. A triple $p(s,o) \in \mathcal{T}$ is a fact consisting of subject $s$, relation $p$ and object $o$ where $s \in \mathcal{E}$, $p \in \mathcal{R}$, $o \in \mathcal{E}$. From a logical point of view, a relation is a binary predicate and the entities in $\mathcal{E}$ are constants. Figure~\ref{KG} shows a small example of a knowledge graph. The entities $\mathcal{E}$ described in this graph are cities, counties, countries and currencies. Some of these entities are linked with each other via one of the three relations in $\mathcal{R}$.

The term \textit{link prediction} origins from calling a relation between two entities a link. In a realistic link prediction application, we have an incomplete knowledge graph $\mathcal{T}$ and we use a link prediction model to create new triples $\mathcal{T'}$ with $\mathcal{T'} \cap \mathcal{T} = \emptyset$. If the approach works well, most of the triples in $\mathcal{T'}$ are correct even though they have been missing in $\mathcal{T}$. Within an evaluation context we split a given knowledge graph $\mathcal{T}$ into train set $\mathcal{T}_{train}$, test set $\mathcal{T}_{test}$ and a validation set $\mathcal{T}_{valid}$. This split is jointly exhaustive and pairwise disjoint. $\mathcal{T}_{train}$ is usually used to train a model, $\mathcal{T}_{valid}$ can be used to optimize the hyperparameters and $\mathcal{T}_{test}$ to evaluate the model performance. Link prediction, often interchangeably called knowledge graph completion, is the task of predicting a target entity given a source entity and a relation. We call such a task a completion task. Each test triple $t_{test} \in \mathcal{T}_{test}$ results in two completion tasks $p(s,?)$ and $p(?,o)$. The task of link prediction is to predict the correct candidate that acts as a substitution for the $?$ in the query.


In \textbf{transductive} link prediction the test set $\mathcal{T}_{test}$ only contains test triples where subject and object appear in the training set $\mathcal{T}_{train}$. Given a triple $p(s,o) \in \mathcal{T}_{test}$, we have at least one triple in train where $s$ appears in subject or object position and at least one triple where $o$ appears in subject or object position. The information encoded in the training set serves two purposes. It is used to train the model and it contains also the information that describes the relevant features. 

\begin{figure}
\caption{Difference between transductive (on the left) and inductive link prediction (on the right).}
\medskip
\centering
\scalebox{1.0}{
\begin{tikzpicture}[]
\scriptsize
\begin{scope}[every node/.style={}]
    \node (g1) at (-0.5,-1.8) {Train/Test graph};
    \node (g2) at (4.0, -1.8) {Train graph};
    \node (g2) at (7.25, -1.8) {Test graph};
\end{scope}



\begin{scope}[shift={(-1.3,0.8)}]
\draw[fill=violet!50] (0,0) circle (0.75cm);
\draw[pattern={crosshatch},pattern color=teal!50] (0,0) circle (0.75cm);
\draw[fill=blue!70] (0,0) --  (40:0.75) arc(40:80:0.75) -- cycle;
\draw[fill=red!70] (0,0) --  (0:0.75) arc(0:40:0.75) -- cycle;
\draw[color=white, rotate=3, postaction={decorate, decoration={text along path, raise=1pt,  ,text align={align=right}, text={|\tiny|training + inference \ \ \ valid \ test}, reverse path}}] (0,0) circle (0.8cm);
\end{scope}

\begin{scope}[shift={(0.5,0)}, every node/.style={circle,draw, inner sep=2pt, fill=white}]
\draw
(-0.5,0.9) node(a1){} (0.0,1.1) node(a2){} (0.5,1.0) node(a3){}
(-0.75,0.5) node(b1){} (-0.25,0.5) node(b2){} (0.25,0.4) node(b3){} (0.75,0.5) node(b4){}
(-1.0,0.0) node(c1){} (-0.5,0.1) node(c2){} (0.0,0.0) node(c3){} (0.5,-0.2) node(c4){} (1.0,0.0) node(c5){}
(-0.75,-0.5) node(d1){} (-0.25,-0.5) node(d2){} (0.2,-0.4) node(d3){} (0.75,-0.5) node(d4){}
(-0.6,-1.0) node(e1){} (0.0,-1.2) node(e2){} (0.6,-1.0) node(e3){};


\draw[violet!50, line width=2pt] (b1) -- (a1) -- (b2) -- (a2) -- (b3) -- (a3) -- (b4);
\draw[teal!50, line width=2pt, dotted] (b1) -- (a1) -- (b2) -- (a2) -- (b3) -- (a3) -- (b4);
\draw[violet!50, line width=2pt] (e1) -- (d1) -- (e2) -- (d2) -- (e2) -- (d3) -- (e2);
\draw[teal!50, line width=2pt, dotted] (e1) -- (d1) -- (e2) -- (d2) -- (e2) -- (d3) -- (e2);
\draw[teal!50, line width=2pt] (b1) -- (c1) -- (c2) -- (c3) -- (b3) -- (c4) -- (c5);
\draw[violet!50, line width=2pt, dotted] (b1) -- (c1) -- (c2) -- (c3) -- (b3) -- (c4) -- (c5);
\draw[teal!50, line width=2pt] (e3) -- (d4) -- (c5);
\draw[violet!50, line width=2pt, dotted] (e3) -- (d4) -- (c5);
\draw[teal!50, line width=2pt] (c1) -- (d1);
\draw[violet!50, line width=2pt, dotted] (c1) -- (d1);
\draw[blue!70, line width=2pt] (b4) -- (c4);
\draw[blue!70, line width=2pt] (e1) -- (e2);
\draw[blue!70, line width=2pt] (b1) -- (c2);
\draw[red!70, line width=2pt] (c1) -- (d1);
\draw[red!70, line width=2pt] (a2) -- (a3);
\draw[red!70, line width=2pt] (d3) -- (d4);

\end{scope}

\draw[line width=1pt] (2,1.75) -- (2,-2.0) ;


\begin{scope}[shift={(3.5,0.8)}]
\draw[fill=teal!50] (0,0) circle (0.75cm);
\draw[fill=blue!70] (0,0) --  (0:0.75) arc(0:40:0.75) -- cycle;
\draw[color=white, rotate=3, postaction={decorate, decoration={text along path, raise=1pt,  ,text align={align=right}, text={|\tiny|training \ \ \ valid}, reverse path}}] (0,0) circle (0.8cm);
\end{scope}

\begin{scope}[shift={(4.75,-0.7)}, every node/.style={circle,draw, inner sep=2pt, fill=white}]
\draw
(-0.5,0.9) node(a1){} (0.0,1.1) node(a2){} (0.5,1.0) node(a3){}
(-0.75,0.5) node(b1){} (-0.25,0.5) node(b2){} (0.25,0.4) node(b3){} (0.75,0.5) node(b4){}
(-1.0,0.0) node(c1){} (-0.5,0.1) node(c2){} (0.0,0.0) node(c3){} (0.5,-0.2) node(c4){};

\draw[teal!50, line width=2pt] (a1) -- (b2) -- (a2) -- (b3) -- (a3) -- (b4);
\draw[teal!50, line width=2pt] (b1) -- (c1) -- (c2) -- (c3) -- (b3);
\draw[teal!50, line width=2pt] (b4) -- (c4);
\draw[teal!50, line width=2pt] (b1) -- (c2);
\draw[teal!50, line width=2pt] (a2) -- (a3);
\draw[blue!50, line width=2pt] (b1) -- (a1);
\draw[blue!50, line width=2pt] (b3) -- (c4);

\end{scope}

\begin{scope}[shift={(6.75,0.9)}]
\draw[fill=violet!50] (0,0) circle (0.75cm);
\draw[fill=red!70] (0,0) --  (0:0.75) arc(0:40:0.75) -- cycle;
\draw[color=white, rotate=3, postaction={decorate, decoration={text along path, raise=1pt,  ,text align={align=right}, text={|\tiny|inference \ \ \ test}, reverse path}}] (0,0) circle (0.8cm);
\end{scope}

\begin{scope}[shift={(7.5,0.4)}, every node/.style={circle,draw, inner sep=2pt, fill=white}]
\draw
(1.0,0.0) node(c5){}
(-0.75,-0.5) node(d1){} (-0.25,-0.5) node(d2){} (0.2,-0.4) node(d3){} (0.75,-0.5) node(d4){}
(-0.6,-1.0) node(e1){} (0.0,-1.2) node(e2){} (0.6,-1.0) node(e3){};
\draw[violet!50, line width=2pt] (e1) -- (d1) -- (e2) -- (d2) -- (e2) -- (d3) -- (e2);
\draw[violet!50, line width=2pt] (e1) -- (d1) -- (e2) -- (d2) -- (e2) -- (d3) -- (e2);
\draw[violet!50, line width=2pt] (e3) -- (d4) -- (c5);
\draw[violet!50, line width=2pt] (e1) -- (e2);
\draw[violet!50, line width=2pt] (d3) -- (d4);
\draw[red!70, line width=2pt] (d1) -- (e2) -- (d3);

\end{scope}

\end{tikzpicture}
}
\label{fig:indeval}
\end{figure}
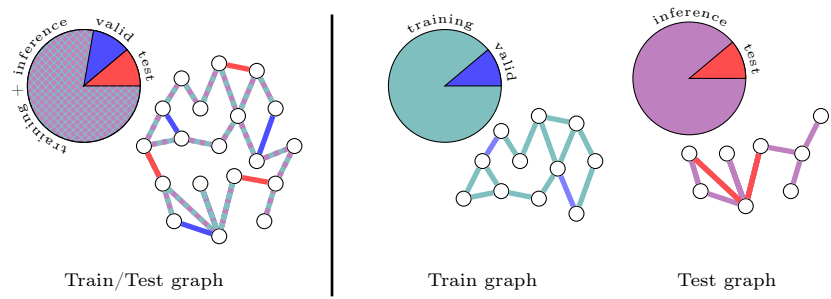

We illustrated this setting on the left side of Figure~\ref{fig:indeval}. The area in the pie-chart colored with a violet/teal grid refers to $\mathcal{T}_{train}$, the blue area to $\mathcal{T}_{valid}$ and the red area to $\mathcal{T}_{test}$. Below the pie chart, we depicted an example graph. We used the same coloring scheme for the edges of the triples that belong to the respective sets. The important aspect of the transductive setting is related to the fact that each of the entities involved in a test triple is described also via one or several triples in the train graph. The fact that we have this type of overlap between $\mathcal{T}_{train}$ and $\mathcal{T}_{test}$ is a characteristic feature of the transductive setting, which allows, for example, to use embeddings of entities learned in the training phase to score candidates in the test phase.

On the right side of Figure~\ref{fig:indeval} we depicted the \textbf{inductive} setting. 
Here we have a strict separation of entities that appear in the graph used to train the model, and entities that appear in the triples of the test set. Note that the test graph contains not not only the test triples (red), but also triples (purple) that describe the entities that appear in the test triples. In the context of inductive link predictions this set of triples is sometimes called inference set, as it is required to infer the missing triples of the test set. The entities described in the training graph are not overlapping with the entities used in the test graph.


It is, nevertheless, possible to learn a model on a set of entities that are disjoint from the entities to which the model is applied. This is the case because both sets of entities are described by the same set of relations. If we would learn a rule-based model, we might for example learn the rule $genre(X,Y) \leftarrow prequel(X,Z), genre(Z,Y)$. This rule says that a movie has the same genre (with a certain probability) as its prequel (if there is a prequel of that movie). 
It is obviously applicable in the transductive and inductive setting. However, a rule containing constants such as $nationality(X,dutch) \leftarrow bornIn(X,amsterdam)$ (a person born in Amsterdam has nationality Dutch) can only be applied to the transductive setting.


\subsection{Evaluation Protocol}
 \label{sub:evalprotocol}

The most prominent metrics for measuring the quality of link prediction methods are ranking-based metrics. They are calculated based on the position of the correct candidate within a ranking of candidate entities. These candidates are elements of a set of possible entities $\mathcal{E^*}$. The target candidate itself is always an element of $\mathcal{E^*}$. In the transductive setting, we have usually $\mathcal{E^*} = \mathcal{E}$. This means that all known entities have to be ranked. In the inductive setting, $\mathcal{E^*}$ is usually a sample from $\mathcal{E}$ with $|\mathcal{E^*}| \ll |\mathcal{E}|$. We call an evaluation protocol which asks to rank all possible candidates a \textbf{non-sampling evaluation protocol} contrary to a protocol that asks to rank the candidates from a restrictive set, which we call a \textbf{random sampling evaluation protocol}.

Furthermore, according to the filtering protocol of~\cite{bordes2013translating}, already known triples in the knowledge graph are ignored when calculating the ranking. This filtered setting is the quasi standard and unfiltered scores are rarely reported. We follow this approach and report always filtered scores in our experiments.

The most prominent metrics are the hits@k and the mean reciprocal rank (MRR). Let $\mathcal{I}$ be the set of ranks of the correct candidates for the completion tasks $p(s,?)$ and $p(?,o)$ derived from all test triples $p(s,o)$ in a given datasets. 


Hits@k is then the ratio of test queries where the rank $i$ of the correct candidate is less than or equal to $k$ and MRR is the arithmetic mean over individual reciprocal ranks.
Let $\mathcal{I}$ be the set of ranks $i$ of the correct candidates for the completion tasks $p(s,?)$ and $p(?,o)$ derived from all test triples $p(s,o)$ in a given datasets. Then we have:
\begin{align*}
    hits@k = \frac{1}{|\mathcal{I}|} \sum_{i \in \mathcal{I}}^{} 1 \text{ if } i \leq k   \ \ \ \ \ \ MRR = \frac{1}{|\mathcal{I}|} \sum_{i \in \mathcal{I}}^{} \frac{1}{i}
\end{align*}


We argue in the following that a small and, in particular, randomly chosen $\mathcal{E^*}$ makes its easy to achieve high hits@10 and MRR scores. This is especially the case if the completion methods under discussion is tailored to such a setting, while such a method will perform poor in any realistic setting.

Sometimes classification metrics as AUC-PR are reported~\cite{grail,indigo}. To calculate the AUC-PR, a single negative triple is constructed for each test triple $p(s,o)$ by randomly corrupting the head or the tail with a random entity. 
Since we want to compare our results with those of the transductive setting where primarily ranking-based metrics are used, we will only focus on them in the following. Nevertheless, our findings are also applicable to the AUC-PR metric and similar metrics.

\section{The Random Sampling Evaluation Protocol}


In~\cite{grail} the authors proposed inductive link prediction as research problem and introduced a pioneering approach called GraIL. In this paper the authors also introduced the most commonly used benchmarks. These benchmarks would support also a non-sampling based evaluation protocol. However, the authors decided to use a random sampling-based variant of the evaluation protocol due to performance issues of GraIL. Instead of ranking $\mathcal{E}$ for each completion task they used a randomly chosen sample $\mathcal{E^*}$ of 50 candidates (one of them is the correct target candidate). Unfortunately most succeeding work~\cite{nodepiece,nbfnet,snri,conglr,compile,indigo} adapted this evaluation protocol. Thus, the random sampling approach has become the defacto standard for evaluating inductive link prediction.


Using the random sampling evaluation protocol the set of sampled negative entities contains a variety of entities of different types. Due to this variety, the type of the sampled negative often does not make sense. For example, for the completion task $genre(Star Wars, ?)$ the set of negatives could contain entities such as countries, artists or professions. Due to the lack of type-matched negatives it is sufficient for an approach to simply make predictions about the type of an entity rather than the entity itself to achieve high scores. 

Furthermore, all approaches evaluated using this evaluation protocol report hits@10. This high choice of $k$ further limits the ability to evaluate the models real inductive link prediction performance, as a model which randomly assigns scores to entities on average already achieves a hits@10 score of 0.2 as the probability of assigning a top-10 score to the correct entity by chance is $10/50=0.2$. 

\subsection{Baseline}

We propose a simple baseline in which candidates are simply ranked higher if they are valid answers to a prediction task depending on the entity type of the candidate. For each relation $p$ we learn rules that determine whether entities are valid candidates for the subject position and for the object position of $p$. We iterate over all possible pairs of relations $h$ and $b$ ($h \neq b$). If the set of subject entities of relation $b$ has a high overlap with the set of subject entities of $h$, the presence of an outgoing relation $b$ of an entity can be used to predict the outgoing relation $h$. We thus learn rules adhering to the following language bias
 \begin{align*}
 r_{ss}: \ \ & h(X,A) \leftarrow b(X,B) &  r_{so}: \ \ & h(X,A) \leftarrow b(B,X)\\
 r_{os}: \ \ & h(A,X) \leftarrow b(X,B) &  r_{oo}: \ \ & h(A,X) \leftarrow b(B,X)
 \end{align*}
\noindent where $h$ is the head relation and $b$ is the relation of the body atom. Uppercase letters represent variables. $A$ and $B$ are unbound variables. Note that we learn rules for any combination of subject and object positions of relations $h$ and $b$.

These rules are an indirect way to implicitly infer the type of an entity. Here a type expresses the validity of entities for the subject/object position of a relation. For each rule we calculate a confidence which can be used as how well a rule predicts an entity $X$ for being a valid type for the subject or object position of $h$ depending on whether the bound variable $X$ is on the subject or object position of the rule. This confidence is the relative proportion of correctly predicted subject/object positions of all predicted entities $conf(r) = |\{x : x \in X_b \land x \in X_h\}|/|\{x : x \in X_b\}|$ where $X_b$ is the set of subjects/objects (depending on the position of the bound variable $X$) of triples in the training set having body relation $b$, while $X_h$ is the set of subjects/objects of triples in the training set having head relation $h$.
 
Given a list of entities to be ranked as predictions for a prediction task $p(s,?)$, we then rank those entities higher that based on the confidence of the rule have a more suitable type for $?$. For each head $p(?,o)$ and tail $p(s,?)$ query in the test set of the test graph we apply the rules learned by our baseline. Since the same entity can be proposed by multiple rules, we order the entities via the confidence of the rule with the highest confidence. 

For example, from Figure~\ref{KG} we might learn a rule $r_1$ that says that an entity might be a good candidate for the subject of the relation $currency$ if it is the object of a triple with the relation $capital$. As both require entities of type \textit{country} this rule makes sense. Furthermore, subjects of relation $locatedIn$ can take on a mixture of entities of type \textit{county} and \textit{city}. A rule such as $r_2$ makes sense, however it can only predict a subset of types. Due to the mixture of types of subjects of relation $locatedIn$, its ability to determine entities of type \textit{city} based on its presence is rather limited. Thus the confidence of $r_3$ is rather low.
\begin{align*} 
r_1: \ \ & currency(X,A) \leftarrow capital(B,X) & \ 1.0 [3/3] \\
r_2: \ \ & locatedIn(X,A) \leftarrow capital(X,B) & \ 1.0 [3/3] \\
r_3: \ \ & capital(X,A) \leftarrow locatedIn(X,B) & \ 0.3 [3/10]
\end{align*} 

This baseline should result in low inductive link prediction performance as it is simply predicting linked entities based on correct types. For example a rule $marriedTo(X,A) \leftarrow bornIn(B,X)$ could be learned, which would have almost a perfect confidence as every person that was married was also born somewhere. In the context of link prediction this does not make sense as this would mean that e.g. the link prediction query $marriedTo(?,paul)$ would be answered with every person born. However, as we will show in later experiments, when evaluated using the flawed random sampling evaluation protocol this baseline appears to perform very well in inductive link prediction.

\subsection{Type-Matched Sampling Protocol}

Approaches which require repeated sampling of subgraphs such as GraIL~\cite{grail} and its variants~\cite{conglr,compile,snri} are very slow to evaluate. In fact the flawed sampling evaluation protocol was introduced to help overcome this limitation. In order to be able to evaluate approaches where it is infeasible to score and rank all entities we release a collection of type-matched negatives (TMN) for each benchmark dataset. For each test query we provide 50 negatives that adhere to the type of the searched answer and thus create a more challenging set of negatives. For example for the test triple {\fontfamily{qcr}\selectfont<Indie rock, /music/genre/artists, Oasis>} of FB15k-237 v3 the set of type-matched head negatives contains genre entities like {\fontfamily{qcr}\selectfont Pop rock, Hip hop music} or {\fontfamily{qcr}\selectfont Death metal} while the set of tail negatives contains artist entities like {\fontfamily{qcr}\selectfont Katy Perry, The Smashing Pumpkins, Miley Cyrus} or {\fontfamily{qcr}\selectfont AC/DC}.

We created these sets by applying the rules learned by our baseline for each head $p(?,o)$ and tail $p(s,?)$ query in the test set as described above. Since we aggregate the confidences of an entity predicted by multiple rules using the maximum it could be possible that one rule dominates the creation of the 50 negatives. For example if the rule with the highest confidence would be $city(X,A) \leftarrow capital(X,B)$ and would predict more than 50 entities the resulting set of negatives would only contain capitals.
To circumvent this and provide a good mixture of high confident entities we bin the predictions in three buckets: entities with a confidence $c \geq 0.75$, entities with a confidence $0.75 > c \geq 0.25$ and entities with a confidence $0.25 > c$. First we try to randomly sample 50 negative entities from the bucket of entities with the highest confidence. If this bucket contains less than 50 entities, we try to sample missing entities from the bucket with the next highest confidence and so on until the a set of negatives reaches a size of 50 entities. If the baseline does not generate enough entities we fill missing entities with randomly sampled entities. Furthermore in each step we keep only true negative entities, which means that the resulting negative triples only contain triples that do not appear in the knowledge graph. This is true for all benchmarks and versions except NELL-995 v1 where due to the limited size of entities it was not possible to create 50 non-positive candidates for all queries. As mentioned in the introduction, we publicly release sets of type-matched negatives for all the benchmarks used in this paper, and code to generate such sets for other datasets.

\section{Experimental Evaluation}

\subsection{Datasets, Approaches and Metrics}

For the evaluation of different approaches we use the commonly used inductive link prediction benchmarks created by \cite{grail}. They were created by sampling two disjunct subgraphs from the transductive benchmark datasets FB15k-237~\cite{toutanova2015observed}, WN18RR~\cite{dettmers2018convolutional} and NELL-995~\cite{nell}. Each benchmark was released as 4 size-increasing different versions denoted v1, v2, v3 and v4. 
Table~\ref{tab:stats} shows the statistics of the benchmarks. It should be noted that the relation counts differ for unknown reasons from the original statistics in~\cite{grail}. 

\begin{table}
\scriptsize
\centering
\caption{Number of relations (\#R), entities (\#E) and training, validation and test triples of the training and test graph for each benchmark dataset.}
\begin{tabular}{lc|ccccc|ccccc|ccccc}
\toprule
& \multirow{2}{*}{Graph} & \multicolumn{5}{c|}{FB15k-237} & \multicolumn{5}{c|}{WN18RR} & \multicolumn{5}{c}{NELL-995} \\
& & \#R & \#E & Train & Valid & Test & \#R & \#E & Train & Valid & Test & \#R & \#E & Train & Valid & Test\\

\midrule
\multirow{2}{*}{v1} & train &  \multirow{2}{*}{180} & 1594 & 4245 & 489 & 492 &  \multirow{2}{*}{9} & 2746 & 5410 & 630 & 638 &  \multirow{2}{*}{14} & 3103 & 4687 & 414 & 439 \\
 & test & & 1093 & 1993 & 206 & 205 &  & 922 & 1618 & 185 & 188 & & 225 & 833 & 101 & 100 \\
\midrule
 \multirow{2}{*}{v2} & train &  \multirow{2}{*}{200} & 2608 & 9739 & 1166 & 1180 &  \multirow{2}{*}{10} & 6954 & 15262 & 1838 & 1868 &  \multirow{2}{*}{88} & 2564 & 8219 & 922 & 968 \\
 & test & & 1660 & 4145 & 469 & 478 & & 2757 & 4011 & 411 & 441 & & 2086 & 4586 & 459 & 476 \\
\midrule
 \multirow{2}{*}{v3} & train & 215 & 3668 & 17986 & 2194 & 2214 &  \multirow{2}{*}{11} & 12078 & 25901 & 3097 & 3152 &  \multirow{2}{*}{142} & 4647 & 16393 & 1851 & 1873 \\
 & test & & 2501 & 7406 & 866 & 865 & & 5084 & 6327 & 538 & 605 & & 3566 & 8048 & 811 & 809 \\
\midrule
 \multirow{2}{*}{v4} & train &  \multirow{2}{*}{219} & 4707 & 27203 & 3352 & 3361 &  \multirow{2}{*}{9} & 3861 & 7940 & 934 & 968&  \multirow{2}{*}{76} & 2092 & 7546 & 876 & 867 \\
 & test & & 3051 & 11714 & 1416 & 1424 & & 7084 & 12334 & 1394 & 1429 & & 2795 & 7073 & 716 & 731 \\
\bottomrule
 \end{tabular}
\label{tab:stats}
\end{table}

We perform experiments using different approaches for inductive link prediction, which can be categorized in rule-based (Neural-LP~\cite{neurallp}, DRUM~\cite{drum}, RuleN~\cite{rulen}, AnyBURL~\cite{meilicke2019anyburl}) and graph neural network (GNN) based (GraIL~\cite{grail}, ConGLR~\cite{conglr}, COMPILE~\cite{compile}, SNRI~\cite{snri}, INDIGO~\cite{indigo}, NBFNet~\cite{nbfnet}, AStarNet~\cite{astarnet}, RED-GNN~\cite{redgnn}). An exception to this classification is NodePiece~\cite{nodepiece} which can be used without a GNN in theory. However, due to its low performance otherwise, it is used in combination with a GNN in the context of inductive link prediction. 
We will now give a short overview of the different approaches. Approaches described within the same bullet point belong to the same family of approaches. 
\begin{itemize}
\item RuleN~\cite{rulen} and AnyBURL~\cite{meilicke2019anyburl} are both symbolic rule-based approaches, which extract first-order logic rules in the form of horn clauses directly from knowledge graphs. Both calculate an approximated rule confidence which is used to score predictions. 
\item  NeuralLP~\cite{neurallp} and DRUM~\cite{drum} are end-to-end differentiable approaches for rule learning. NeuralLP uses TensorLog operations to learn first-order logic rules, while DRUM uses bidirectional RNNs. 
\item NBFNet~\cite{nbfnet}, RED-GNN~\cite{redgnn} and  AStarNet~\cite{astarnet} are both approaches that progressively propagate from a node to its neighborhood in a breadth-first-searching (BFS) manner. RED-GNN is utilizing dynamic programming for efficient subgraph encoding, while NBFNet is based on the Bellman-Ford algorithm for shortest path problems. AStarNet enhances scalability by learning to preselect important nodes and edges.
\item Usually entities are encoded using shallow encoding which means there is one embedding for each entity. NodePiece~\cite{nodepiece} is an approach to reduce the number of embeddings needed by embedding-based approaches. The embedding of a node is retrieved using its relational context and the distance of it to selected anchor nodes. In the inductive setting only the relational context is used in combination with a relational message passing GNN based on CompGCN~\cite{vashishth}.
\item INDIGO~\cite{indigo} is another GNN-based approach, which encodes the triples of the knowledge graph pair-wise, i.e. nodes in the graph correspond to pairs of entities. It is trained as a denoising autoencoder.
\item GraIL~\cite{grail} is a GCN-based approach adapted to the inductive link prediction setting. It scores triples consisting of unseen entities by performing message passing on extracted enclosing subgraphs between the target nodes using k-hop neighbourhood. Nodes in extracted subgraphs are labelled with respect to the target nodes using a double radius vertex labeling scheme. GraIL and all its extensions require to extract a subgraph for each link which is not scalable to large graphs. This was the main reason for using a random sampling evaluation protocol.
\item CoMPILE, ConGLR and SNRI extend GraIL by adding additional processing steps. CoMPILE~\cite{compile} introduces the ability to naturally handle asymmetric/antisymmetric relations. ConGLR~\cite{conglr} can be seen as a hybrid approach between GCN-based and rule-based approaches. It extends GraIL by the additional extraction of a context graph from the k-hop neighbourhood subgraph. SNRI~\cite{snri} extends GraIL by not only considering the enclosing subgraph between the target node, but also partial neighbouring relations by applying mutual information (MI) maximization.
\end{itemize}

We train each approach using the settings from their respective papers. We refer to \url{https://github.com/nomisto/inductiveeval} for details. Due to the amount of different versions of benchmark datasets we will mainly discuss averages over the four versions for each dataset. For better comparability with the random sampling protocol, for which hits@10 results have often been reported, we will also base most of our discussion on hits@10. Tables containing the results using both hits@10 and MRR can be found at the URL above. 

\subsection{Results}

\subsubsection{Random Sampling Evaluation Protocol}

The upper part of Table~\ref{tab:composite} labeled \textit{Random Sampling} shows the hits@10 results using the random sampling evaluation protocol. If available, we report the numbers presented in the original paper. If not available, we report results based on our own experiments with these approaches. As a cross-check, we compare all our models with models for which numbers are available. We observed only insignificant differences from the numbers of the original papers.

Our  simple baseline outperforms current state-of-the-art approaches on FB15k-237. It achieves an average hits@10 of about 0.938 over the different versions by simply distinguishing the correct entity based on its type. The baseline also outperforms or rivals the performance of state-of-the-art approaches on NELL-995, especially on the bigger versions V3 and V4. Since WN18RR is a knowledge graph of word senses and general relations such as \textit{hypernym}, it does not contain strongly typed entities. As a result, the baseline fails to achieve competitive results with the exception of the V3 version. The reason for that are different distributions of relations of the test sets. V3 is dominated by relations for which the tail entity is always an entity from a small set of entities (for example countries or regions), which can be learned by the baseline.


These results illustrate already clearly that the random sampling evaluation protocol together with the hits@10 metric is not an appropriate protocol to evaluate and rank approaches for inductive link prediction. With respect to the average hits@10 score our baseline ranks at position \#1, \#13 and \#3 out of 14 approaches (including the baseline) on the three evaluation datasets. We also measured the MRR results, to check if we can observe a similar behaviour. For the MRR it is more important to rank the correct candidate at the top position, thus, the MRR has a stronger focus on a high precision. Here our baselines occupies the positions \#7, \#13 and \#13. While the MRR solves some issues of the random sampling protocol, the baseline still achieves a midfield position for FB15k-237 and there are 7 approaches that perform worse.


\subsubsection{Non-Sampling Evaluation Protocol} 
The middle part of Table~\ref{tab:composite} labeled \textit{Non-Sampling} shows the hits@10 results for the non-sampling evaluation protocol, i.e. ranking the correct candidate within all entities of the knowledge graph. Since the evaluation of GraIL and its extensions is very inefficient, we decided to only evaluate benchmarks where the evaluation does not take significantly longer than 12 hours. For example the evaluation of GraIL on FB15k-237 V3 takes 10 hours, while the evaluation of CoMPILE/ConGLR/SNRI on FB15k-237 V2 already takes 9/10.75/11.5 hours.

The non-sampling evaluation approach draws a completely different picture. Similar to the random sampling evaluation protocol, NBFNet still can be considered as the best performing approach by achieving the highest average hits@10 on FB15k-237 and NELL-995, while achieving the second highest average hits@10 on WN18RR. NodePiece, which showed one of the best performances when evaluated using the random sampling evaluation protocol, loses over-proportionally as it now only scores 0.450 average hits@10, scoring 0.476 percentage points lower compared to the random sampling evaluation protocol, while NBFNet only scores 0.257 percentage points lower. NodePiece, which was the best performing approach given the results of the random sampling protocol, has been overtaken by five other approaches in the realistic non-sampling protocol. One reason for its good performance on the random sampling evaluation could be, that as NodePiece primarily uses 1-hop incoming and outgoing relations of a node, it could be prone to leverage typing information of a node. AStarNet and RED-GNN appear to behave similarly to NBFNet, with a difference in scores of about 0.221 percentage points for AStarNet and 0.207 percentage points for RED-GNN compared to the random sampling evaluation protocol.
{\tabcolsep 1.0pt

\begin{table}
\scriptsize
\centering
\caption{Hits@10 results using the \textbf{random sampling} (upper part), \textbf{non-sampling} (middle part) and \textbf{type-matched sampling} evaluation protocol (bottom part).
$\dagger$ results were generated by us, $^\star$ results are from~\cite{grail}, other results for \textbf{random sampling} evaluation protocol are from the original papers. All results using the \textbf{non-sampling} (middle part) and \textbf{type-matched sampling} evaluation protocol were generated by us.
To compensate the variance of the random sampling protocol, we report the average of 100 evaluation runs.
RED-GNN~\cite{redgnn} and AStarNet~\cite{astarnet} also report results using the non-sampling evaluation protocol in their respective papers. However, for reasons unknown, they do not only evaluate on the test set of the test graph, but rather a combination of the test set of the test graph and a further hold-out set of the test graph. The usage of this hold-out set, sometimes referred to as validation set of the test graph, is not really clear and is neither used for training, validation or testing in the original paper~\cite{grail}. Numbers reported by us were created using solely the test set however.
}
\resizebox{\textwidth}{!}{%
\begin{tabular}{llccccr|ccccr|ccccr}
\toprule
& \multirow{2}{*}{Approach} &  \multicolumn{5}{c}{FB15k-237} & \multicolumn{5}{c}{WN18RR} & \multicolumn{5}{c}{NELL-995} \\
\cmidrule(lr{.15em}){3-7} \cmidrule(lr{.15em}){8-12} \cmidrule(lr{.15em}){13-17}
& & V1 & V2 & V3 & V4 & avg & V1 & V2 & V3 & V4 & avg & V1 & V2 & V3 & V4 & avg  \\
\midrule
\parbox[t]{3mm}{\multirow{14}{*}{\rotatebox[origin=c]{90}{Random Sampling}}} & Neural LP$^\star$ & .529 & .589 & .529 & .559 & .552 & .744 & .689 & .462 & .671 & .642 & .408 & .787 & .827 & .806 & .707 \\
& DRUM$^\star$ & .529 & .587 & .529 & .559 & .551 & .744 & .689 & .462 & .671 & .642 & .194 & .786 & .827 & .806 & .653 \\
& RuleN$^\star$ & .498 & .778 & .877 & .856 & .752 & .809 & .782 & .534 & .716 & .710 & .535 & .818 & .773 & .614 & .685 \\
& AnyBURL$^\dagger$ & .517 & .784 & .845 & .865 & .753 & .814 & .776 & .544 & .719 & .713 & .795 & .824 & .775 & .605 & .750 \\
\cmidrule{2-17}
& GraIL$^\star$ & .642 & .818 & .828 & .893 & .795 & .825 & .787 & .584 & .734 & .733 & .595 & .933 & .914 & .732 & .794 \\
& ConGLR~\cite{conglr} & .683 & .860 & .886 & .893 & .831 & .856 & \textbf{.929} & .707 & \textbf{.929} & \underline{.855} & .811 & \underline{.949} & .944 & .816 & .880 \\
& COMPILE~\cite{compile} & .676 & .830 & .847 & .874 & .807 & .836 & .798 & .607 & .755 & .749 & .584 & .939 & .928 & .752 & .801 \\
& SNRI~\cite{snri} & .718 & .865 & .896 & .894 & .843 & .872 & .831 & .673 & .883 & .815 & .643$^\dagger$ & .900$^\dagger$ & .918$^\dagger$ & .191$^\dagger$ & .663$^\dagger$ \\
& INDIGO$^\dagger$  & .451 & .480 & .483 & .480 & .473 & .166 & .156 & .316 & .203 & .210 & .520 & .552 & .510 & .497 & .520 \\
& NodePiece~\cite{nodepiece} & \underline{.873} & .939 & .944 & \underline{.949} & \underline{.926} & .830 & .886 & \underline{.785} & .807 & .827 & .890 & .901 & .936 & .893 & \underline{.905} \\
& NBFNet~\cite{nbfnet} & .834 & \underline{.949} & \underline{.951} & \textbf{.960} & .924 & \textbf{.948} & \underline{.905} & \textbf{.893} & \underline{.890} & \textbf{.909} & \underline{.908}$^\dagger$ & \textbf{.956}$^\dagger$ & \textbf{.968}$^\dagger$ & \textbf{.939}$^\dagger$  & \textbf{.943}$^\dagger$ \\
& AStarNet$^\dagger$ & .687 & .867 & .904 & .916 & .844 & \underline{.877} & .836 & .715 & .799 & .807 & \textbf{.940} & .899 & .860 & .828 & .882 \\
& RED-GNN$^\dagger$ & .516 & .771 & .829 & .903 & .755 & .856 & .824 & .684 & .777 & .785 & .853 & .925 & .954 & .806 & .884 \\
\cmidrule{2-17}
& Baseline$^\dagger$ & \textbf{.887} & \textbf{.963} & \textbf{.954} & \underline{.949} & \textbf{.938} & .087 & .145 & .590 & .138 & .240 & .761 & .912 & \underline{.945} & \underline{.914} & .883 \\
\bottomrule
\toprule
\parbox[t]{3mm}{\multirow{13}{*}{\rotatebox[origin=c]{90}{Non-Sampling}}} & NeuralLP & .351 & .437 & .388 & .341 & .379 & .713 & .684 & .312 & .636 & .586 & .750 & .598 & .544 & .585 & .619 \\
& DRUM & .385 & .468 & .410 & .403 & .417 & .715 & .680 & .310 & .635 & .585 & .755 & .554 & .585 & \underline{.591} & .621 \\
& AnyBURL & .461 & .640 & .620 & .614 & .584 & .803 & .772 & .467 & \underline{.705} & .687 & .830 & .610 & .486 & .530 & .614 \\
\cmidrule{2-17}
& GraIL  & .376 & .415 & .423 & \textit{t/o} & \textit{.404} & .729 & .744 & .406 & .675 & .638 & .685 & .310 & .336 & .219 & .387 \\
& ConGLR & .373 & .423 & \textit{t/o} & \textit{t/o} & \textit{.398} & .566 & .688 & \textit{t/o}  & \textit{t/o} & \textit{.627} & .810 & .512 & \textit{t/o} & \textit{t/o} & \textit{.661} \\
& COMPILE & .417 & .498 & \textit{t/o} & \textit{t/o} & \textit{.457} & .689 & .656 & \textit{t/o} & \textit{t/o} & \textit{.673} & .760 & .606 & \textit{t/o} & \textit{t/o} & \textit{.683} \\
& SNRI & .246 & .416 & \textit{t/o} & \textit{t/o} & \textit{.331} & .319 & .465 & \textit{t/o} & \textit{t/o} & \textit{.392} & .800 & .407 & \textit{t/o} & \textit{t/o} & \textit{.603}\\
& INDIGO & .198 & .247 & .246 & .198 & .222 & .040 & .011 & .056 & .005 & .028 & .615 & .230 & .193 & .182 & .305 \\
& NodePiece & .429 & .500 & .451 & .419 & .450 & .540 & .661 & .262 & .506 & .492 & .895 & .294 & .306 & .178 & .418 \\
& NBFNet & \textbf{.617} & \textbf{.719} & \textbf{.659} & \textbf{.661} & \textbf{.664} & \textbf{.830} & \underline{.787} & \textbf{.554} & .700 & \underline{.718} & \underline{.925} & \textbf{.694} & \textbf{.688} & \textbf{.640} & \textbf{.737} \\
& AStarNet & \underline{.534} & \underline{.673} & \underline{.647} & \underline{.637} & \underline{.623} & \underline{.819} & \textbf{.804} & \underline{.541} & \textbf{.736} & \textbf{.725} & \textbf{.960} & \underline{.690} & \underline{.646} & .440 & \underline{.684} \\
& RED-GNN  & .420 & .590 & .582 & .600 & .548 & .774 & .776 & .532 & .696 & .694 & .840 & .582 & .553 & .301 & .569 \\
\cmidrule{2-17}
& Baseline & .351 & .384 & .253 & .211 & .300 & .005 & .002 & .022 & .000 & .007 & .835 & .124 & .129 & .049 & .284 \\
\bottomrule
\toprule
\parbox[t]{3mm}{\multirow{12}{*}{\rotatebox[origin=c]{90}{Type-matched Sampling}}} & NeuralLP & .441 & .550 & .499 & .531 & .505 & .729 & .689 & .336 & .652 & .601 & .750 & .731 & .734 & .748 & .741 \\
& DRUM & .446 & .556 & .497 & .533 & .508 & .729 & .689 & .336 & .652 & .601 & .755 & .745 & .752 & .758 & .752 \\
& AnyBURL & .495 & .712 & .717 & .746 & .668 & .814 & .776 & .540 & .719 & .712 & .830 & .759 & .645 & .586 & .705 \\
\cmidrule{2-17}
& GraIL & .566 & .690 & .688 & .546 & .622 & .840 & .810 & .596 & .762 & .752 & .720 & .798 & .719 & .613 & .713 \\
& ConGLR & \underline{.685} & .659 & .694 & .699 & .684 & .785 & .804 & .501 & .753 & .711 & .865 & \textbf{.889} & \underline{.853} & .769 & \underline{.844} \\
& COMPILE & .622 & \underline{.759} & .735 & .724 & .710 & .835 & .800 & .591 & .746 & .743 & .765 & \underline{.888} & .838 & .574 & .766 \\
& SNRI & .554 & .697 & .679 & .696 & .656 & \underline{.878} & .815 & .595 & .773 & .765 & .805 & .776 & .663 & .371 & .654 \\
& INDIGO  & .344 & .383 & .388 & .391 & .376 & .136 & .091 & .183 & .131 & .135 & .615 & .478 & .420 & .418 & .483 \\
& NodePiece & .532 & .615 & .625 & .600 & .593 & .798 & \underline{.842} & .599 & .765 & .751 & .905 & .555 & .575 & .470 & .626 \\
& NBFNet & \textbf{.688} & \textbf{.794} & \textbf{.776} & \textbf{.784} & \textbf{.760} & \textbf{.941} & \textbf{.910} & \textbf{.863} & \textbf{.899} & \textbf{.903} & \underline{.970} & .861 & \textbf{.868} & \textbf{.819} & \textbf{.880} \\
& AStarNet & .651 & .753 & \underline{.754} & \underline{.761} & \underline{.730} & .875 & .832 & \underline{.702} & \underline{.797} & \underline{.802} & \textbf{.975} & .828 & .794 & \underline{.775} & .843 \\
& RED-GNN  & .483 & .700 & .718 & .754 & .664 & .854 & .823 & .665 & .777 & .780 & .895 & .842 & .807 & .540 & .771 \\
\bottomrule
 \end{tabular}
}
\label{tab:composite}
\end{table}}

The predictive performances of rule-based approaches, such as the symbolic rule-based approach AnyBURL, seem to be actually better than assessed using the random sampling evaluation protocol. AnyBURL has an average hits@10 of 0.584 on FB15k-237, only 0.169 percentage points less than the average sampled hits@10, achieving competitive results comparable to NBFNet. On WN18RR AnyBURL scores an average hits@10 of about 0.687, where the best performing approaches NBFNet achieve 0.664 and AStarNet 0.623. The difference of sampling and non-sampling average hits@10 for DRUM also only amount to 0.134 percentage points and for NeuralLP 0.173. 

\subsubsection{Type-matched Sampling Evaluation Protocol}

Using sampled negatives that have an appropriate type draws a similar picture than using the non-sampling protocol. However, this protocol has the advantage that it is also applicable to less efficient models. The bottom part of Table~\ref{tab:composite} labeled \textit{Type-matched Sampling} shows the hits@10 using the type-matched sampling evaluation protocol. This evaluation protocol still shows that NBFNet is one of the top-performing approaches as it achieves the best average hits@10 on all datasets. However approaches, such as NodePiece, that performed well on the random sampling evaluation approach as they might learn types of entities, do not perform well in this setting. It is still evident that rule-based approaches produce competitive results. Thus, the type-matched sampling protocol can be used as substitute for the non-sampling protocol in case of scalability issues.

\subsubsection{Comparing different protocols}

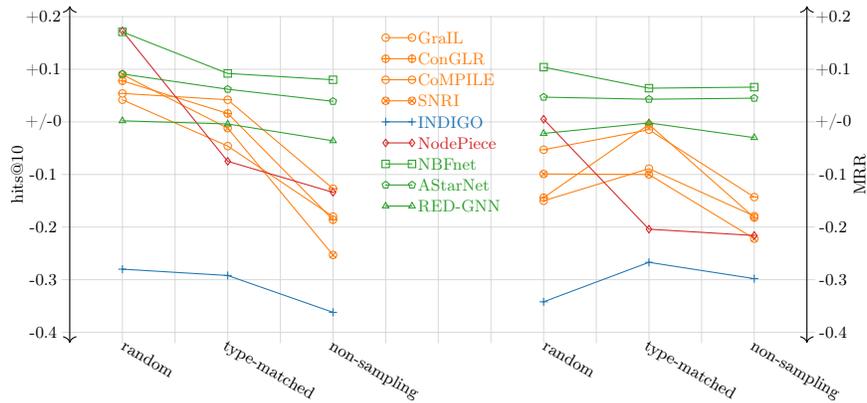
\begin{figure}
    \caption{Absolute changes in performance of different approaches compared to AnyBURL under different evaluation protocols (random sampling, type-matched and non-sampling) using average hits@10 (left) and average MRR (right) on FB15k-237.}
    \medskip
    \centering
    \scalebox{1.0}{
    \begin{tikzpicture}[scale=0.7, every node/.style={scale=0.7}, cap=round]
\draw[step=1cm,very thin,color=black!15] (-0.15,-4.2) grid (14.15,2);
\draw (1.6,-4.3) node[anchor=north,rotate=-30] {random};
\draw (3.9,-4.5) node[anchor=north,rotate=-30] {type-matched};
\draw (5.9,-4.5) node[anchor=north,rotate=-30] {non-sampling};
\draw (9.6,-4.3) node[anchor=north,rotate=-30] {random};
\draw (11.9,-4.5) node[anchor=north,rotate=-30] {type-matched};
\draw (13.9,-4.5) node[anchor=north,rotate=-30] {non-sampling};
\draw[<->] (0,-4.2) -- (0,2.2);
\draw[<->] (14,-4.2) -- (14,2.2);

\draw (-1.0,-1.0) node[rotate=90] {hits@10};
\draw (-0.5,-4) node {-0.4};
\draw (-0.5,-3) node {-0.3};
\draw (-0.5,-2) node {-0.2};
\draw (-0.5,-1) node {-0.1};
\draw (-0.5,0) node {+/-0};
\draw (-0.5,1) node {+0.1};
\draw (-0.5,2) node {+0.2};

\draw (15.0,-1.0) node[rotate=90] {MRR};
\draw (14.5,-4) node {-0.4};
\draw (14.5,-3) node {-0.3};
\draw (14.5,-2) node {-0.2};
\draw (14.5,-1) node {-0.1};
\draw (14.5,0) node {+/-0};
\draw (14.5,1) node {+0.1};
\draw (14.5,2) node {+0.2};

\definecolor{mycolor_red}{rgb}{0.8392156862745098, 0.15294117647058825, 0.1568627450980392}
\definecolor{mycolor_blue}{rgb}{0.12156862745098039, 0.4666666666666667, 0.7058823529411765}
\definecolor{mycolor_orange}{rgb}{1.0, 0.4980392156862745, 0.054901960784313725}
\definecolor{mycolor_violet}{rgb}{0.17254901960784313, 0.6274509803921569, 0.17254901960784313}

\draw[mycolor_orange, mark=o] plot coordinates{ (1.0,0.42) (3.0,-0.46) (5.0,-1.8) };
\draw[mycolor_orange, mark=oplus] plot coordinates{ (1.0,0.78) (3.0,0.16) (5.0,-1.86) };
\draw[mycolor_orange, mark=halfcircle] plot coordinates{ (1.0,0.54) (3.0,0.419999999999999) (5.0,-1.27) };
\draw[mycolor_orange, mark=otimes] plot coordinates{ (1.0,0.9) (3.0,-0.12) (5.0,-2.53) };
\draw[mycolor_blue, mark=+] plot coordinates{ (1.0,-2.8) (3.0,-2.92) (5.0,-3.62) };
\draw[mycolor_red, mark=diamond] plot coordinates{ (1.0,1.73) (3.0,-0.750000000000001) (5.0,-1.34) };
\draw[mycolor_violet, mark=square] plot coordinates{ (1.0,1.71) (3.0,0.92) (5.0,0.800000000000001) };
\draw[mycolor_violet, mark=pentagon] plot coordinates{ (1.0,0.91) (3.0,0.619999999999999) (5.0,0.39) };
\draw[mycolor_violet, mark=triangle] plot coordinates{ (1.0,0.02) (3.0,-0.04) (5.0,-0.359999999999999) };

\draw[mycolor_orange, mark=o] plot coordinates{ (6.0,1.6) (6.5, 1.6) } node[anchor=west] {GraIL};
\draw[mycolor_orange, mark=oplus] plot coordinates{  (6.0,1.2) (6.5, 1.2) }node[anchor=west] {ConGLR};
\draw[mycolor_orange, mark=halfcircle] plot coordinates{  (6.0,0.8) (6.5, 0.8) } node[anchor=west] {CoMPILE};
\draw[mycolor_orange, mark=otimes] plot coordinates{  (6.0,0.4) (6.5, 0.4) }node[anchor=west] {SNRI};
\draw[mycolor_blue, mark=+] plot coordinates{ (6.0,-0.0) (6.5, -0.0) } node[anchor=west] {INDIGO};
\draw[mycolor_red, mark=diamond] plot coordinates{ (6.0,-0.4) (6.5, -0.4) } node[anchor=west] {NodePiece};
\draw[mycolor_violet, mark=square] plot coordinates{ (6.0,-0.8) (6.5, -0.8) } node[anchor=west] {NBFnet};
\draw[mycolor_violet, mark=pentagon] plot coordinates{ (6.0,-1.2) (6.5, -1.2) } node[anchor=west] {AStarNet};
\draw[mycolor_violet, mark=triangle] plot coordinates{ (6.0,-1.6) (6.5, -1.6) } node[anchor=west] {RED-GNN};

\draw[mycolor_orange, mark=o] plot coordinates{ (9.0,-1.5) (11.0,-0.89) (13.0,-1.79) };
\draw[mycolor_orange, mark=oplus] plot coordinates{(9.0,-1.44) (11.0,-0.05) (13.0,-1.82) };
\draw[mycolor_orange, mark=halfcircle] plot coordinates{ (9.0,-0.53) (11.0,-0.15) (13.0,-1.43) };
\draw[mycolor_orange, mark=otimes] plot coordinates{ (9.0,-0.990000000000001) (11.0,-1) (13.0,-2.22) };
\draw[mycolor_blue, mark=+] plot coordinates{ (9.0,-3.42) (11.0,-2.67) (13.0,-2.98) };
\draw[mycolor_red, mark=diamond] plot coordinates{ (9.0,0.05) (11.0,-2.04) (13.0,-2.16) };
\draw[mycolor_violet, mark=square] plot coordinates{ (9.0,1.04) (11.0,0.639999999999999) (13.0,0.66) };
\draw[mycolor_violet, mark=pentagon] plot coordinates{ (9.0,0.469999999999999) (11.0,0.43) (13.0,0.45) };
\draw[mycolor_violet, mark=triangle] plot coordinates{ (9.0,-0.22) (11.0,-0.02) (13.0,-0.3) };

\end{tikzpicture}
    }
    \label{fig:traj}
\end{figure}

Figure~\ref{fig:traj} shows trajectories that illustrate how the performance of the different approaches changes on the FB15k-237 benchmark, when evaluated using the random, type-matched and non-sampling evaluation protocols. As the different protocols get increasingly difficult we normalize the performance of each approach by showing the absolute difference to the performance of AnyBURL. As can be seen by the green trajectories, robust approaches such as NBFNet, AStarNet and RED-GNN can maintain a steady advantage compared to AnyBURL. NodePiece, however, which scored 0.173 better on hits@10 than AnyBURL evaluated using the random sampling protocol, scores about 0.068 lower on hits@10 than AnyBURL when evaluated under type-matched sampling protocol and substantially lower (0.134) when evaluated under non-sampling protocol. 
This might be a hint that this approach learns types very well instead of learning the likelihood of links between entities. GraIL and its extensions show similar behaviour using hits@10, where the lead of these approaches is diminishing when evaluated under type-matched sampling protocol and lose lead by a margin when evaluated under non-sampling protocol. On MRR AnyBURL already performs better than GraIL and its extensions using the random sampling protocol. Using the type-matched sampling protocol, they could make up leeway and perform similar or only a little worse than AnyBURL. Using the non-sampling protocol, MRR behaves similar to hits@10 again.

\section{Conclusion}

Current inductive link prediction approaches are evaluated by comparing a prediction only against a small set of randomly selected negatives. Due to the limited size of the set of negatives, errors are introduced that prevent the assessment of the model's true inductive link prediction ability. We show that current state-of-the-art results for inductive link prediction can be achieved using a simple rule-based baseline. Furthermore, we re-evaluate approaches to inductive link prediction, where the order of the state-of-the-art changes drastically. Our corrected results indicate that GNN-based approaches such as NBNnet, AStarNet and RED-GNN are clearly ahead compared to the other approaches. The rule-based approach AnyBURL is an efficient and fully explainable alternative that performs only slightly worse. At the same time we observed that approaches as NodePiece and GraIL as well as its successors lag behind if we evaluate them under a realistic protocol.

We advocate that further research should be evaluated using an evaluation protocol where all entities of the knowledge graph are used as negatives. If it is not possible to evaluate an approach on all entities, we publish datasets of type-matched head and tail negatives and encourage inductive link prediction researchers to use them instead of randomly sampling negatives.

\begin{credits}
\subsubsection{\ackname} This work was funded by the Austrian security research program KIRAS of the Federal Ministry of Finance (BMF) through the  DAGMAR project (grant No. 52224305).

\subsubsection{\discintname}
The authors have no competing interests to declare that are
relevant to the content of this article.
\end{credits}
%
%
%

\bibliographystyle{splncs04}
\bibliography{references}

\end{document}